\documentclass[10pt,twocolumn,letterpaper]{article}

\usepackage{wacv}    
\usepackage[accsupp]{axessibility} %

\usepackage{multirow}
\usepackage{booktabs}

\definecolor{wacvblue}{rgb}{0.21,0.49,0.74}
\usepackage[pagebackref,breaklinks,colorlinks,allcolors=wacvblue]{hyperref}

\def\wacvPaperID{16} %
\def\confName{WACV}
\def\confYear{2027}

\title{ Auditing Liver Fibrosis Staging under Scanner Shift}

\author{Asmita Yuki Pritha\\
Capstone School Dhaka\\
Dhaka, Bangladesh\\
{\tt\small asmitayukipritha@gmail.com}
\and
Xin Wang\\
University at Albany, SUNY\\
Albany, NY, USA\\
{\tt\small xwang56@albany.edu}
\and
Shu Hu\thanks{Corresponding author.}\\
Purdue University\\
West Lafayette, IN, USA\\
{\tt\small shuhu@purdue.edu}
}

\begin{document}
\maketitle

\begin{abstract}
Pooled cross validation, in which every scanner appears in every training fold, is a common way to evaluate liver fibrosis staging models. When case mix differs between scanners, it can reward a model for recognizing the scanner rather than the liver. We audit a staging pipeline with a deep and a radiomic stream on 460 MRI examinations from three scanners, comparing protocols that differ only in which scanners contribute labeled training data; the primary endpoint, a within scanner C index, was fixed in advance. Pooled cross validation concealed an extraction error that emptied the liver crop in a quarter of the examinations. We found it by tracing why the forest's cirrhosis AUC on scanner B2 was 0.726 under pooled folds but 0.171 with B2 held out. Under leave one scanner out evaluation, where one scanner had guided earlier development, no declared alternative, including three planned interventions, beat the deep stream (C index 0.646), and the radiomic forest, which was indistinguishable from it in distribution, was significantly worse. Every representation we tested encoded the scanner, and label free harmonization neither removed this signal nor lifted the radiomic stream clearly above the deep stream. A histopathology replication with three frozen foundation models showed inflated accuracy under splits that share patients (slight for the two pathology specific encoders), near perfect hospital identification, and no transfer benefit from harmonization. Here, the choice of evaluation protocol changed which pipeline looked best. Code: {\urlstyle{same}\href{https://github.com/Purdue-M2/auditing-liver-fibrosis-staging}{\nolinkurl{github.com/Purdue-M2/auditing-liver-fibrosis-staging}}}.
\end{abstract}

\section{Introduction}
\label{sec:intro}
 
Chronic liver disease causes about two million deaths each year~\cite{asrani2019burden}, and in most patients it progresses through fibrosis, whose stage guides surveillance and treatment. Biopsy is still the reference standard, but it is invasive and samples only a tiny part of the organ. Biopsies taken from the two lobes of the same liver often differ by at least one stage~\cite{regev2002sampling}. Imaging offers a noninvasive alternative. Deep networks have staged fibrosis from gadoxetic acid enhanced MRI~\cite{yasaka2018liver,hectors2021fully} and contrast enhanced CT~\cite{choi2018development} with promising accuracy, and radiomic pipelines built on liver shape and texture features have been proposed for multi series liver MRI~\cite{zhou2026}.
 
Most of this work rests on an assumption that is usually left implicit: \emph{that accuracy measured on new patients from the training scanners will hold on a new scanner}. The assumption breaks down when scanner and disease are entangled, as they are here. In the recently released LiFS data~\cite{lifs}, cirrhosis accounts for 38\% of the examinations on scanner A, 61\% on B1 and 52\% on B2, and B2 has only five examinations without substantial fibrosis. Knowing which scanner produced an examination therefore already says something about its stage. A model can exploit this, and pooled cross validation, which places every scanner in every training fold, is poorly placed to notice. Worse, it can hide an error whose footprint differs by scanner, as happened in our own pipeline.
 
The danger is well documented outside liver imaging. Chest radiograph models have learned the hospital along with the disease~\cite{zech2018variable,degrave2021ai}, histology models encode the submitting site~\cite{howard2021impact}, and systematic reviews have traced many optimistic results to flawed validation~\cite{roberts2021common,varoquaux2022machine,kapoor2023leakage}. The usual remedies are harmonization~\cite{johnson2007combat,fortin2018harmonization} and domain generalization~\cite{ganin2015unsupervised}. Both are useful, but their benefit can only be judged on a scanner that played no part in development, and controlled comparisons suggest that it is often smaller than reported~\cite{gulrajani2021search}. Before asking how to remove scanner effects, it is worth asking what actually transfers to a new scanner, and what only appears to.
 
Our work takes a different direction. Rather than propose another invariance method, we audit a complete two stream staging pipeline, a ConvNeXt deep stream on the last contrast enhanced phase and a radiomic random forest on seven MR series, under protocols that differ only in which scanners contribute labeled training data. We fixed the primary endpoint, a within scanner concordance index that uses every patient, before computing it for any model, and we account for every planned comparison. Leave one scanner out evaluation, endpoints fixed in advance and direct measurement of scanner information each have precedents, but to our knowledge they have not been combined in fibrosis staging.
 
We make four contributions. \textbf{(1)} We show that pooled cross validation concealed an extraction error that emptied the liver crop in 115 of 460 examinations. The radiomic forest reached a cirrhosis AUC of 0.726 on B2 under pooled folds but only 0.171 with B2 held out, and tracing that gap is what exposed the error. In a controlled simulation, pooled evaluation also barely registers injected errors of this kind. \textbf{(2)} Under leave one scanner out evaluation, where one scanner had guided earlier development, no declared alternative, including three planned interventions, beats the deep stream on the primary endpoint (C index 0.646, 95\% CI 0.602 to 0.688). The radiomic forest is significantly worse ($-0.085$, $p = 0.003$), fusion leads in mean AUC only because of one scanner and task pair with five negatives, and on scanner A a frozen DINOv2 probe does better. We present it with the caveat that B2 guided the development of the deep recipe. \textbf{(3)} We find that every representation we tested encodes the scanner (AUC 0.857 to 0.998), that the strength of this signal does not predict transfer, and that label free harmonization neither removes this signal nor lets the radiomic stream clearly overtake the deep stream. \textbf{(4)} On CAMELYON17 histology patches, three frozen foundation models show the same three effects: inflated accuracy when patches of one patient appear in both training and testing (small for the two pathology specific encoders), encoded hospitals, and harmonization that does not improve transfer. Since the gap between the same models changes with the scanners they are trained on, we argue that in multi scanner staging the evaluation protocol is part of the method.

\section{Related Work}
\label{sec:related}

\paragraph{Fibrosis staging from imaging.}
Noninvasive staging has followed two routes. Deep networks learn directly from MRI~\cite{yasaka2018liver,hectors2021fully} or CT~\cite{choi2018development}. Radiomic pipelines instead pass handcrafted shape, intensity and texture descriptors~\cite{haralick1973textural,lambin2017radiomics,zwanenburg2020ibsi} to a conventional classifier. Zhou \etal~\cite{zhou2026}, for example, computed shape, texture, appearance and direction features from multiple MR series, and our radiomic stream adapts this design. These studies report accuracy on internal or external test sets. To our knowledge, none of them treats each scanner in turn as an unseen test domain, compares deep and radiomic pipelines under that protocol with an endpoint fixed in advance, or measures how much scanner information each representation retains.

\paragraph{Site effects, shortcuts and validation design.}
There is strong evidence that imaging models exploit site specific signals. A pneumonia detector could identify the hospital system and generalized worse to external sites~\cite{zech2018variable}, COVID-19 radiograph models selected shortcuts over signal~\cite{degrave2021ai}, and site signatures inflated histology accuracy when sites were shared between training and testing~\cite{howard2021impact}. Geirhos \etal~\cite{geirhos2020shortcut} called this behavior shortcut learning, and Castro \etal~\cite{castro2020causality} argued that it should be analyzed causally. Closest to our analysis, scanner effects in multi site brain MRI survived harmonization~\cite{glocker2019machine}. Work on protected attributes adds that a model can encode an attribute without relying on it, and that reducing the encoding does not guarantee fair transfer~\cite{glocker2023algorithmic,brown2023detecting,yang2024limits}; we therefore measure scanner decodability and transfer separately. Pooled cross validation across scanners is not leakage in the strict sense~\cite{kapoor2023leakage}, since no test label enters training, but it places the test scanner in every training fold, and our extraction error shows what that overlap can hide.

\paragraph{Harmonization and robust training.}
Remedies act at different levels. ComBat~\cite{johnson2007combat} harmonizes the location and scale of each feature per batch and is widely used for multi center imaging data~\cite{fortin2018harmonization,orlhac2022guide}. Domain adversarial training~\cite{ganin2015unsupervised} removes source information from the representation, whereas Pritha \etal~\cite{pritha2026robust} train a multi source CT classifier to also predict its source center, with a logit adjusted auxiliary loss so that small centers are not overlooked. Rank based losses weight samples by the rank of their loss~\cite{fan2017topk,hu2020learning,hu2023rank}, and our deep stream uses one of them~\cite{fan2017topk}; Li \etal~\cite{li2026robust} combine logit adjustment with group level CVaR for fair CT diagnosis. Whether any remedy helps on a new scanner depends on how it is evaluated. Under fair model selection, domain generalization methods rarely beat empirical risk minimization~\cite{gulrajani2021search,koh2021wilds}, which motivates both our protocol and our caveat that the deep recipe was selected with one scanner in view.

\paragraph{Frozen representations and prevalence shift.}
Frozen foundation models are often proposed as a way around site variation, from CLIP features for COVID-19 detection in CT~\cite{lin2024robust} to pathology encoders such as UNI~\cite{chen2024uni}, Phikon~\cite{filiot2023phikon} and Phikon-v2~\cite{filiot2024phikonv2}, although site differences measurably shift learned representations~\cite{stacke2021measuring}. We therefore evaluate frozen DINOv2, Phikon and Phikon-v2 embeddings as well. Yet even a representation from which the scanner cannot be recovered would not fix the decision threshold, because a change in prevalence alone moves it~\cite{lipton2018detecting}, and calibration degrades under dataset shift more generally~\cite{ovadia2019trust}. We therefore report threshold transfer separately from ranking performance.

\section{Methods}
\label{sec:methods}
 
\begin{figure*}[t]
\centering
\includegraphics[width=\linewidth]{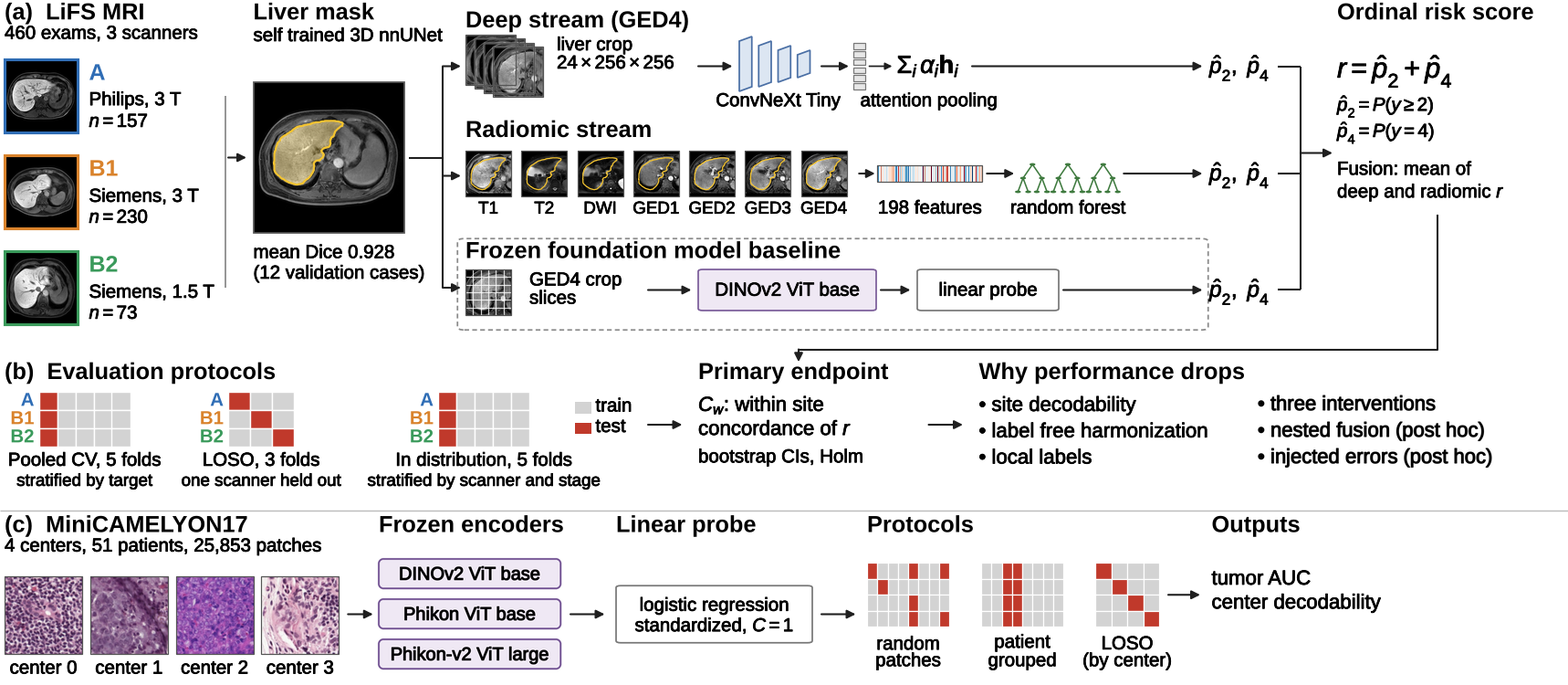}
\caption{\textbf{Study design.} (a) Two staging streams and a frozen DINOv2 baseline, each reduced to the ordinal risk score $r$ (Sec.~\ref{sec:data} and \ref{sec:models}). (b) Evaluation protocols, primary endpoint and the analyses of why performance drops (Sec.~\ref{sec:protocols} and \ref{sec:why}). (c) The protocol comparison repeated on histopathology (Sec.~\ref{sec:histo}). Images are real examples from LiFS and MiniCAMELYON17.}
\label{fig:method}
\end{figure*}
 
Our study has three parts. We first evaluate one staging pipeline, with a deep and a radiomic stream, under protocols that differ only in how sites are split between training and testing. We then ask why performance drops at a new site, and finally repeat the protocol comparison in histopathology. Throughout, a \emph{site} means a scanner in MRI and a center in histopathology. Figure~\ref{fig:method} summarizes the design, and Table~\ref{tab:design} shows which analyses were fixed before we saw their results.
 
\subsection{Data and liver masks}
\label{sec:data}
 
We use the 460 training cases of the LiFS dataset~\cite{lifs}, acquired on three scanners: A (Philips Ingenia, 3\,T; $n=157$), B1 (Siemens Skyra, 3\,T; $n=230$) and B2 (Siemens Aera, 1.5\,T; $n=73$). Each case comes with one fibrosis stage, S1 to S4, and up to seven series: T1 and T2 weighted images, diffusion weighted imaging at $b=800$\,s/mm$^2$ and four contrast enhanced series, GED1 to GED4. Some series are unavailable, for instance T2 in 7 cases and diffusion in 35, but GED4 is present in every case. Following the dataset, we predict two binary targets, substantial fibrosis (S2 or higher) and cirrhosis (S4). The case mix differs strongly between scanners. S4 accounts for 38\%, 61\% and 52\% of the cases on A, B1 and B2, and S1 for 35\%, 17\% and 7\%, so B2 has only five cases without substantial fibrosis, a fact that shaped our choice of endpoint.
 
Manual liver masks exist for only 30 cases, ten per scanner, all on GED4. We kept 12 of them (four per scanner) for validation, trained a 3D nnUNet~\cite{isensee2021nnunet} with the large residual encoder preset~\cite{isensee2024revisited} on the remaining 18, and refined it with two rounds of self training on the 430 unmasked volumes, keeping only confident pseudo labels. The final weights reach a mean Dice of 0.928 on the 12 validation cases. Two caveats apply: the same cases guided checkpoint selection, so the estimate may be optimistic, and one of them stays below a Dice of 0.73 with each network we checked. The deep crops use the selected checkpoint, while most radiomic features use the final weights. Because the masks were not regenerated for each fold, the segmentation model has seen images and liver masks from every scanner, although never a fibrosis label.
 
\subsection{Staging models}
\label{sec:models}
 
\textbf{Deep stream.} The deep stream reads GED4 only. The liver bounding box, enlarged by 10\% on each side, is cropped, resampled to $24\times256\times256$ voxels and standardized. A ConvNeXt Tiny encoder~\cite{liu2022convnext} initialized from ImageNet embeds each of the 24 slices, with its first convolution summed over the RGB filters to accept a single channel. Attention pooling~\cite{ilse2018attention} combines the slice embeddings, and a linear layer with dropout of 0.5 predicts the four stages. The loss is a class weighted cross entropy in which only the largest 80\% of per sample losses in a batch are averaged, that is, the average top $k$ loss of~\cite{fan2017topk} with $k=3$ for our batches of four. Training runs for 40 epochs with AdamW (learning rate $10^{-4}$, weight decay $10^{-5}$), a cosine schedule and random in plane flips. No checkpoint is selected: the final weights are used, and predictions are averaged over four flips and three seeds.
 
\textbf{Radiomic stream.} The radiomic stream uses all seven series. Each series is resampled onto a $64\times128\times128$ grid around the padded mask and standardized inside the mask, and from each we extract 10 first order, 10 Haralick texture~\cite{haralick1973textural} and 6 gradient statistics. With 16 shape and intensity distribution features, this yields 198 features; a 33 feature version with GED4 and shape features only serves as a control. For each target we train a random forest~\cite{breiman2001random} with 600 trees, a minimum leaf size of 2 and balanced class weights, after imputing missing values with the training median. To test whether the choice of classifier matters, we also fit logistic regression ($C=0.1$), support vector machines with linear and radial basis kernels, and gradient boosting to the same features.
 
\textbf{Risk score and fusion.} Every model returns two probabilities, $\hat p_2=P(y\geq 2)$ and $\hat p_4=P(y=4)$, for stage $y\in\{1,\dots,4\}$; for the deep stream, $\hat p_2=1-P(y=1)$. Their sum is our ordinal risk score,
\begin{equation}
r=\hat p_2+\hat p_4\in[0,2],
\label{eq:risk}
\end{equation}
so S4 cases, which meet both conditions, should score highest. Fusion is the unweighted mean of the deep and radiomic risk scores. We did not tune the weights, so no test scanner can influence them. As a foundation model baseline, each crop slice ($224\times224$, three channels) passes through a frozen DINOv2 ViT base~\cite{oquab2024dinov2}, the slice embeddings are averaged, and a logistic regression probe ($C=0.1$) is fit on top.
 
\subsection{Protocols and statistics}
\label{sec:protocols}
 
We compare three protocols. \emph{Pooled cross validation}, which our development pipeline used for the radiomic stream, splits all 460 cases into five folds stratified by target, so every scanner appears in every training fold. \emph{Leave one site out} (LOSO) trains on two scanners and tests once on the third. In the declared LOSO analyses, nothing is selected or tuned on the held out scanner and its labels are never used; the two post hoc selections and the local label experiment are marked as exceptions. As an \emph{in distribution} reference, five fold cross validation stratified by scanner and stage, with the same three seeds for the deep stream as under LOSO, shows what each model reaches when the test scanner is represented in training.
 
One caveat applies to every LOSO result. The deep and radiomic recipes were developed earlier on these 460 cases, trained on A and B1 and checked out of distribution on B2, and are reused here unchanged. No held out scanner is therefore fully untouched, and B2, which served as that check, is the least clean of the three. Baseline settings were fixed and never tuned.
 
\textbf{Primary endpoint.} We decided against an AUC as the primary endpoint for two reasons. B2 has only five S1 cases, so its substantial fibrosis AUC rests on very few negatives, and an AUC pooled over scanners can rise simply because a model separates scanners with different case mix. Before computing it for any model, we therefore fixed a within site concordance,
\begin{equation}
C_w=\frac{\sum_{s}\sum_{(i,j)\in\mathcal{P}_s}\big(\mathbf{1}[r_i>r_j]+\tfrac{1}{2}\,\mathbf{1}[r_i=r_j]\big)}{\sum_{s}|\mathcal{P}_s|},
\label{eq:cw}
\end{equation}
where $\mathcal{P}_s$ is the set of pairs of held out cases from site $s$ with $y_i>y_j$. This is Harrell's concordance~\cite{harrell1982evaluating} over the four ordered stages, counted within sites only. It is 0.5 in expectation for a random score and 1 for a perfect ordering, every case contributes, and a model gains nothing by separating sites. Secondary endpoints are per scanner AUCs, expected calibration error with 10 bins~\cite{guo2017calibration}, and four class accuracy and quadratic weighted kappa~\cite{cohen1968weighted}.
 
\textbf{Inference.} Confidence intervals are 95\% percentile intervals from 2,000 bootstrap resamples~\cite{efron1993bootstrap} drawn within each site; for AUCs, positive and negative cases are resampled separately. Models are compared on the same resamples, the two sided $p$ value is twice the smaller tail of the bootstrap difference, and Holm's procedure~\cite{holm1979simple} controls the familywise error rate at 0.05 within each family of Table~\ref{tab:design}.
 
\begin{table}[t]
\centering
\caption{Analysis plan. An analysis counts as fixed when its endpoint and comparisons were set before any of its results were seen. Post hoc results are exploratory.}
\label{tab:design}
\footnotesize
\begin{tabular}{@{}p{3.45cm}p{2.75cm}p{1.05cm}@{}}
\toprule
Analysis & Status & Test \\
\midrule
Family 1: fusion, forest and DINOv2 vs.\ deep & endpoint fixed before any $C_w$; AUCs seen & Holm (3) \\
Family 2: multiphase, augmentation, adversarial vs.\ deep & fixed before training & Holm (3) \\
ComBat; histopathology arm & fixed before running & none \\
Nested fusion; protocol effect & post hoc & one each \\
All other analyses & descriptive & none \\
\bottomrule
\end{tabular}
\end{table}
 
\subsection{Why performance drops}
\label{sec:why}
 
\textbf{Label free adjustments.} For the radiomic features we test three adjustments aimed at scanner differences: standardization within each scanner, ranks within each scanner, and ComBat~\cite{johnson2007combat,fortin2018harmonization,behdenna2023pycombat} with the scanner as batch, applied to the 184 features with nonzero variance on every scanner. None of them uses labels, but all three use the unlabeled features of the held out scanner, so in practice they require a batch of unlabeled cases from a new scanner before it can be scored.
 
\textbf{Site decodability.} If a representation still identifies the scanner, a classifier trained on it can use scanner differences as a shortcut, so we measure this directly. A random forest with 600 trees predicts the scanner under stratified five fold cross validation, and we report one versus rest AUC and accuracy; always guessing B1, the largest group, gives an accuracy of 0.50. We test the radiomic features under each adjustment, the DINOv2 embeddings, and the pooled embeddings of the deep stream and the three interventions below, all trained with B2 held out (seed 42). Post hoc, we correlate scanner AUC with LOSO C index across the 13 representations for which we have both (Spearman).
 
\textbf{Planned interventions.} Before training them, we fixed three changes to the deep stream, each to be trained with three seeds in every fold and compared with a baseline using the same seeds. The first feeds GED4, T2 and diffusion as three channels resampled onto the crop grid, with a missing series replaced by zeros. The second adds scanner style intensity augmentation (random gamma, bias field, blur and noise)~\cite{zhang2020generalizing}, and the third an adversarial scanner classifier on the pooled embedding, trained through gradient reversal with weight 0.1~\cite{ganin2015unsupervised}. The last two had not finished when this paper was first submitted; we completed them afterwards with the same code.
 
\textbf{Nested fusion (post hoc).} After seeing $C_w$ for each site, we asked whether a fusion could beat the deep stream without using held out labels. For each held out scanner, an inner LOSO on the two training scanners selects one of eight radiomic pipelines (a random forest or a logistic regression, on raw, standardized, ranked or ComBat features), and the selected risk score is averaged with the deep one after both are standardized within the held out scanner.
 
\textbf{Local labels.} Finally, to see what a few labeled cases from a new scanner are worth, we add $k\in\{0,5,10,20,40\}$ of them to radiomic training, repeat this with 10 different draws, and test on the remaining cases of that scanner. 

\textbf{Corruption simulation (post hoc).} To see whether pooled cross validation would also miss other errors like ours, we corrupt the rebuilt radiomic features of 10, 25 or 50\% of one scanner's examinations, chosen at random or with a probability proportional to stage (S4 four times as likely as S1). The corruption zeroes the features of three random series or of all seven, as our error did, or shifts all features by one or two standard deviations, which gives 24 settings. In each, a 300 tree forest is scored under pooled five fold cross validation and with that scanner held out, and the change in C index is averaged over the three scanners and three random draws.
 
\subsection{Histopathology replication}
\label{sec:histo}
 
To see whether these protocol effects are specific to liver MRI, we repeat the comparison on MiniCAMELYON17\footnote{\url{https://huggingface.co/datasets/chehablab/MiniCAMELYON17}, CC0 license.}, a public set of lymph node histology patches labeled for tumor, patient and center. According to its dataset card, it is a tissue filtered sample of $512\times512$ patches at 0.5\,$\mu$m per pixel from part of the CAMELYON17~\cite{bandi2019camelyon17} training slides, drawn with a quota per slide; a patch is tumor when its central third overlaps an annotated metastasis, and ambiguous patches are discarded, as in WILDS~\cite{koh2021wilds}. Four centers contain both classes; together they hold 25,853 patches from 80 slides of 51 patients, 11\% of them tumor. Each patch, resized to $224\times224$, is embedded by three frozen encoders: DINOv2 ViT base, Phikon~\cite{filiot2023phikon}, a ViT base pretrained on histology, and Phikon-v2~\cite{filiot2024phikonv2}, a ViT large pretrained on histology and added post hoc. A standardized logistic regression probe ($C=1$, no tuning) is scored by patch level tumor AUC under random patch splits, patient grouped splits with centers mixed (both five fold), and LOSO, where we again compare no adjustment, per center standardization and ComBat. Center decodability uses the same probe under patient grouped five fold cross validation, with per center standardization fitted on the training folds only, and is compared with a probe that sees no features. Post hoc, we also measure how often a patch's nearest neighbor comes from the same slide, and how well the tumor direction of one slide, its mean tumor minus mean normal embedding, ranks the patches of another patient's slide, using the 11 slides (one per patient) with at least 20 patches of each class. We repeat the analysis on central $224\times224$ crops at native resolution. Confidence intervals resample patients within the held out center (500 resamples), or slides for the tumor direction (2,000 resamples).

\section{Experiments}
\label{sec:experiments}
 
\subsection{Experimental settings}
\label{sec:settings}
 
Beyond \cref{sec:methods}, a few details are needed to read the tables. \Cref{tab:protocol} includes one extra protocol, held out B2, which is simply the B2 fold of leave one scanner out (LOSO) evaluation for the radiomic forest, run once with the original and once with the rebuilt features. Fusion always uses equal weights. Our earlier pipeline had tuned its fusion weights on B2 with pooled forest predictions, but no result in this paper uses those weights, and the forest weight is varied only in a sensitivity analysis (supplementary material). The radiomic stream adapts the shape, texture, appearance and direction features of Zhou \etal~\cite{zhou2026} and reads them with a class balanced forest, while the classical learners, ComBat and the frozen DINOv2 probe are external baselines taken from prior work. Harmonization, nested fusion and the prevalence matched threshold (the quantile of the held out scanner's scores that matches the training prevalence) all use unlabeled data from the held out scanner. They are therefore transductive. Intervals and $p$ values follow \cref{sec:protocols}, with every comparison made on shared resamples. Holm's procedure runs over the three comparisons of each family (\cref{tab:design}). The protocol effect is the forest's C index difference from the deep stream under LOSO minus the same difference in distribution, computed on the same resamples. Learner settings are given in the supplementary material.
 
\begin{table}[t]
\centering
\caption{\textbf{Pooled versus held out evaluation on B2.} Pooled: five fold cross validation over all scanners. Held out: trained on A and B1. Original features contain the extraction error of \cref{sec:results}. Fusion averages the forest with deep predictions from models trained without B2 (AUC 0.571 for S2+, 0.717 for S4), with equal weights. S2+ on B2 has 68 positives and five negatives. $\Delta$ is held out minus pooled.}
\label{tab:protocol}
\small
\setlength{\tabcolsep}{6pt}
\begin{tabular}{lcccc}
\toprule
 & \multicolumn{2}{c}{Forest AUC} & \multicolumn{2}{c}{Fusion AUC} \\
\cmidrule(lr){2-3}\cmidrule(lr){4-5}
Protocol & S2+ & S4 & S2+ & S4 \\
\midrule
\multicolumn{5}{l}{\emph{Original features (198)}} \\
Pooled   & 0.865 & 0.726 & 0.809 & 0.849 \\
Held out & 0.200 & 0.171 & 0.268 & 0.323 \\
$\Delta$ & $-0.665$ & $-0.555$ & $-0.541$ & $-0.526$ \\
\midrule
\multicolumn{5}{l}{\emph{Rebuilt features (198)}} \\
Pooled   & 0.850 & 0.568 & 0.826 & 0.741 \\
Held out & 0.826 & 0.519 & 0.771 & 0.688 \\
$\Delta$ & $-0.024$ & $-0.049$ & $-0.055$ & $-0.053$ \\
\bottomrule
\end{tabular}
\end{table}
 
\begin{table*}[t]
\centering
\caption{\textbf{Primary endpoint under leave one scanner out evaluation.} Pooled within scanner C index of $\hat{P}(\mathrm{S2{+}}) + \hat{P}(\mathrm{S4})$ against the four level stage (95\% bootstrap interval), C index per held out scanner, and difference from the deep stream. Holm decisions apply within each declared family. Post hoc rows belong to no family; the last row's pipeline was chosen with the test results in view. Bold: best declared value per column. The two significant declared comparisons both favor the deep stream.}
\label{tab:primary}
\small
\setlength{\tabcolsep}{4.2pt}
\begin{tabular}{llccccccc}
\toprule
 & Model & C index $\uparrow$ (95\% CI) & A & B1 & B2 & $\Delta$ vs.\ deep (95\% CI) & $p$ & Holm \\
\midrule
Reference & Deep stream (GED4) & \textbf{0.646} (0.602, 0.688) & 0.578 & 0.680 & 0.683 & & & \\
\midrule
\multirow{3}{*}{Family 1} & Equal fusion & 0.625 (0.583, 0.668) & 0.578 & 0.649 & 0.654 & $-0.021$ ($-0.056$, $+0.018$) & 0.248 & n.s. \\
 & Radiomic forest & 0.561 (0.514, 0.606) & 0.575 & 0.555 & 0.544 & $-0.085$ ($-0.139$, $-0.027$) & 0.003 & sig. \\
 & Frozen DINOv2 probe & 0.599 (0.551, 0.643) & \textbf{0.625} & 0.594 & 0.511 & $-0.047$ ($-0.095$, $-0.000$) & 0.048 & n.s. \\
\midrule
\multirow{3}{*}{Family 2} & Multiphase deep input & 0.562 (0.514, 0.609) & 0.482 & 0.616 & 0.482 & $-0.084$ ($-0.136$, $-0.030$) & 0.004 & sig. \\  & Intensity augmentation & 0.609 (0.565, 0.653) & 0.567 & 0.622 & \textbf{0.704} & $-0.037$ ($-0.073$, $+0.002$) & 0.057 & n.s. \\  & Adversarial scanner classifier & 0.640 (0.598, 0.684) & 0.566 & \textbf{0.691} & 0.568 & $-0.006$ ($-0.034$, $+0.023$) & 0.711 & n.s. \\
\midrule
\multirow{3}{*}{Post hoc} & Logistic regression, ComBat & 0.643 (0.598, 0.687) & 0.676 & 0.620 & 0.679 & $-0.003$ ($-0.063$, $+0.055$) & 0.887 & \\
 & Nested fusion & 0.680 (0.640, 0.720) & 0.622 & 0.712 & 0.686 & $+0.034$ ($-0.006$, $+0.073$) & 0.086 & \\
 & Fusion with selected pipeline & 0.687 (0.647, 0.727) & 0.640 & 0.711 & 0.708 & $+0.041$ ($+0.001$, $+0.081$) & 0.046 & \\
\bottomrule
\end{tabular}
\end{table*}

\subsection{Results}
\label{sec:results}

\paragraph{Pooled evaluation concealed an extraction error.}
With the original radiomic features, pooled cross validation gives the forest an AUC of 0.865 for S2+ and 0.726 for S4 on B2. Trained on A and B1 alone, the same forest falls to 0.200 and 0.171 (\cref{tab:protocol}), far below chance. Sampling noise cannot explain a gap this large, and tracing it led us to an extraction error. The crop grid had been anchored at a corner chosen in physical coordinates, and for some image orientations this put the grid outside the liver. In 115 of the 460 examinations (25.0\%) the cropped mask was therefore empty, and the 189 features computed on the crop defaulted to zero. The error stayed hidden because it did not fall evenly on the scanners. Of the 93 affected examinations from A and B1, 74 were cirrhotic, against 200 of all 387 examinations from these two scanners, whereas none of the 22 affected B2 examinations were. A forest trained on A and B1 can thus learn that zeroed crop features mean cirrhosis, a rule that is inverted on B2, and pooled folds can mask the inversion because affected B2 examinations also sit in training. Anchoring the grid at the lowest mask index, along the mask's own axes, removes every empty crop and changes the features of the 345 unaffected examinations very little in the median (relative change 0.0008, 95th percentile 0.116; median rank correlation 0.98). With the grid fixed, the pooled cirrhosis AUC drops from 0.726 to 0.568, so part of the original pooled score came from the error itself. The protocol gap narrows but does not close (\cref{tab:protocol}).

\paragraph{Pooled evaluation also misses injected errors.}
The corruption simulation (\cref{sec:why}) shows the same blindness. In all 24 settings, the forest's pooled C index changed by at most 0.013, whereas its C index on the held out scanner fell by up to 0.090 (shift of two standard deviations) and 0.078 (all seven series zeroed), both with half of the scanner's examinations corrupted in proportion to stage, and the drop grew with the corrupted share. Corruption that followed stage could even raise the pooled C index measured on the corrupted scanner (by up to 0.046), much as the real error had raised the pooled cirrhosis AUC.
 
\paragraph{No declared alternative beats the deep stream under LOSO.}
Under LOSO, the deep stream reaches a pooled C index of 0.646 (95\% CI 0.602 to 0.688), and every declared alternative falls below it (\cref{tab:primary}). The radiomic forest is 0.085 lower ($p = 0.003$), the only family~1 difference that survives Holm correction. The DINOv2 probe is 0.047 lower ($p = 0.048$, not significant after correction), and equal fusion 0.021 lower ($p = 0.248$). For all six declared alternatives, the 95\% interval rules out a gain of more than 0.023. The ranking is not the same on every scanner, and because B2 is the fold most exposed to our development choices, the other two deserve a closer look. On B1, the deep stream leads every family~1 alternative (0.680 against at most 0.649). On A, where it is weakest but still above chance (0.578, 95\% CI 0.51 to 0.65), fusion (0.578) and the forest (0.575) only match it, and DINOv2 (0.625) and all three post hoc models do better. Without the B2 fold, the conclusions do not change, and the per scanner AUCs largely agree (supplementary material). The one large difference in favor of fusion is on B2 S2+ ($+0.200$), the pair with five negatives. That single pair decides the average: across all six pairs fusion appears ahead (0.651 versus 0.627), but without it the order reverses (0.628 versus 0.639). A ranking that hinges on five negatives is fragile, which is why the primary endpoint uses every patient and every stage.
 
In distribution, the comparison looks different. With three deep seeds, as under LOSO, the forest is indistinguishable from the deep stream (C index 0.723 versus 0.750, difference $-0.027$, 95\% CI $-0.066$ to $+0.014$, $p = 0.19$), its pooled AUC is slightly higher (0.790 versus 0.786 for S2+, and 0.785 versus 0.780 for S4), and fusion is best (0.816 and 0.803). A study that stopped at this protocol would therefore have had no reason to prefer the deep stream. Moving to LOSO costs the forest 0.16 in C index and the deep stream 0.10, a protocol effect of $-0.059$ (95\% CI $-0.110$ to $-0.007$, $p = 0.029$); for fusion it is $-0.025$ ($p = 0.141$). On A, every in distribution AUC exceeded 0.80, whereas every LOSO AUC of the two streams lies between 0.568 and 0.612. On B2, the deep stream scores higher held out than in distribution (S4 AUC 0.717 versus 0.587), which fits B2 having guided its development.
 
\paragraph{All three representations encode the scanner.}
A random forest trained with five fold cross validation identifies the scanner from the radiomic features with a one versus rest AUC, macro averaged over the three scanners, of 0.995 and an accuracy of 0.957, against a majority rate of 0.500. The DINOv2 embedding gives 0.998 and 0.970, and the deep embedding 0.935 and 0.839, even though the latter comes from a model trained without B2. Label free harmonization leaves the signal intact, with a scanner AUC of 1.000 after per scanner z scoring and 0.999 after ComBat; aligning the location and scale of each feature presumably leaves enough of the remaining distribution for a forest to work with. Because scanner and stage are correlated, any of them could in principle stage patients by recognizing the scanner, and a held out scanner tests directly whether a model does so. Less scanner information is not the same as better transfer, however: the three interventions lower the deep embedding's scanner AUC to between 0.857 and 0.928, and across 13 representations scanner AUC does not predict LOSO C index (Spearman $\rho = -0.37$, $p = 0.21$).

\paragraph{Classical learners and harmonization do not clearly surpass the deep stream.}
None of the classical learners lifts the radiomic stream clearly above the deep stream (supplementary material). On raw features, mean LOSO AUC ranges from 0.570 (linear SVM) to 0.629 (gradient boosting), against 0.627 for the deep stream. Five of the 17 configurations do edge past the deep stream's mean, by at most 0.011 (logistic regression with ComBat), but we did not test these margins. On the primary endpoint, logistic regression with ComBat reaches a C index of 0.643, on par with the deep stream; because we chose that pipeline after looking at per scanner results, we treat it as an observation rather than a finding. The nested fusion of \cref{sec:why} avoids this selection. The design is post hoc, but no held out label enters any decision; the inner loop chose logistic regression with z scoring when A or B1 was held out, and with raw features when B2 was. Nested fusion reaches 0.680 (95\% CI 0.640 to 0.720), 0.034 above the deep stream, yet the interval of the difference still includes zero ($-0.006$ to $+0.073$, $p = 0.086$). Swapping in the selected pipeline gives 0.687 ($p = 0.046$), a value we list for completeness but do not count.
 
\paragraph{No intervention helps the deep stream.}
The additional series help the forest (supplementary material) but not the deep stream. Supplying GED4, T2 and DWI as three channels (family~2) lowers its C index to 0.562 ($\Delta = -0.084$, 95\% CI $-0.136$ to $-0.030$, $p = 0.004$), the only family~2 difference that survives Holm correction. Intensity augmentation gives 0.609 ($-0.037$, $p = 0.057$) and the adversarial classifier 0.640 ($-0.006$, 95\% CI $-0.034$ to $+0.023$, $p = 0.711$). That weakening the scanner signal does not help is consistent with theory: invariant representations cannot guarantee transfer when the label distribution differs between domains~\cite{zhao2019learning}, as stage does between our scanners.
 
\paragraph{Local labels and decision thresholds.}
Local labels help the forest, most clearly on A and B1 (supplementary material). Fixed decision thresholds transfer poorly. At the default threshold of 0.5, the deep stream's decisions largely reduce to the majority rule. For cirrhosis, its accuracy (0.618 on A, 0.391 on B1, 0.479 on B2) equals that of calling every examination noncirrhotic, and for S2+ on A and B1 it equals that of calling every examination positive. Thresholds tuned on target labels raise cirrhosis accuracy to 0.643, 0.739 and 0.712, and the prevalence matched threshold gives 0.541, 0.574 and 0.671. A model can therefore rank patients better than chance on a new scanner and still classify them poorly at a fixed cutoff. Four class decisions are also uninformative (supplementary material).
 
\begin{table}[t]
\centering
\caption{\textbf{The audit repeated in histopathology.} Tumor versus normal AUC of a fixed logistic probe on frozen embeddings of CAMELYON17 patches from four hospitals. Leave one hospital out (LOHO): mean over held out hospitals. Hospital AUC: one versus rest AUC, macro averaged, of the same probe predicting the hospital under five fold cross validation grouped by patient; with z scoring fitted on the training folds, it equals that of a probe with no features (0.312), which is below 0.5 because grouped folds shift each hospital's share between training and test~\cite{jamalabadi2016classification}. Same slide neighbor: share of patches whose nearest neighbor is from the same slide. Tumor direction: AUC on other patients' slides of the tumor direction of one slide.}
\label{tab:pathology}
\footnotesize
\setlength{\tabcolsep}{3.5pt}
\begin{tabular}{lccc}
\toprule
Protocol or quantity & DINOv2 & Phikon & Phikon-v2 \\
\midrule
Random patch split & 0.998 & 1.000 & 1.000 \\
Folds grouped by patient & 0.849 & 0.990 & 0.985 \\
\midrule
LOHO & 0.877 & 0.977 & 0.957 \\
LOHO, hospital z scoring & 0.872 & 0.929 & 0.925 \\
LOHO, ComBat & 0.876 & 0.933 & 0.921 \\
\midrule
Hospital AUC, raw & 0.978 & 0.999 & 0.999 \\
Hospital AUC, z scoring & 0.311 & 0.309 & 0.314 \\
\midrule
Same slide neighbor & 0.498 & 0.906 & 0.888 \\
Tumor direction & 0.608 & 0.842 & 0.791 \\
\bottomrule
\end{tabular}
\end{table}
 
\paragraph{Replication in histopathology.}
\Cref{tab:pathology} repeats the audit on the histopathology patches of \cref{sec:histo}~\cite{bandi2019camelyon17}, with frozen DINOv2, Phikon and Phikon-v2 embeddings read by an untuned logistic probe, and shows three effects analogous to those in MRI. First, the protocol matters: a random patch split gives DINOv2 an AUC of 0.998, whereas folds grouped by patient give 0.849, although the two pathology encoders change much less (1.000 to 0.990 for Phikon and to 0.985 for Phikon-v2). Second, all three embeddings encode the site: the same probe identifies the hospital with an AUC of 0.978 to 0.999. Third, harmonization does not improve transfer. Per hospital z scoring fitted on the training folds leaves no hospital information the probe can use (hospital AUC 0.309 to 0.314, as for a probe without features), yet LOHO performance stays flat for DINOv2 (0.877 before; 0.872 and 0.876 after z scoring and ComBat) and falls for Phikon (0.977 to 0.929 and 0.933) and Phikon-v2 (0.957 to 0.925 and 0.921).  That the pathology encoders lose so little is not because they ignore the patient: for about nine in ten patches, the nearest neighbor comes from the same slide (0.906 for Phikon and 0.888 for Phikon-v2, against 0.498 for DINOv2). Rather, their tumor signal is shared across patients. The tumor direction of one slide ranks another patient's patches with an AUC of 0.842 and 0.791, against 0.608 for DINOv2 (differences $+0.234$, 95\% CI $+0.159$ to $+0.322$, and $+0.182$, $+0.105$ to $+0.267$). On native resolution crops these findings hold, but the worst hospital and the order of the encoders on single hospitals change, so we do not interpret them.

\section{Conclusion}
\label{sec:conclusion}
 
We audited a two stream liver fibrosis staging pipeline under protocols that differ only in which scanners contribute labeled training data. Pooled cross validation hid the effects of an extraction error affecting a quarter of the examinations; tracing the gap between pooled and scanner held out results is what exposed it. Under a leave one scanner out endpoint fixed in advance, no declared alternative beat the deep stream, whereas the radiomic forest, indistinguishable from it in distribution, fell significantly behind. Every representation we tested encoded the scanner, label free harmonization did not remove this signal, weakening it did not improve transfer, a fixed decision threshold largely collapsed to the majority rule on new scanners, and the same pattern appeared in histopathology. The lesson is not that one stream is better than the other, but that the evaluation protocol changed which one looked better. We therefore recommend reporting leave one scanner out results alongside pooled cross validation, quantifying how well learned representations predict the scanner without reading a lower value as robustness, and evaluating decision thresholds against the class prevalence of each scanner.
 
\paragraph{Limitations.}
Our evidence comes from one dataset with three scanners, and B2 contributes only 73 examinations, five of them without substantial fibrosis. The deep and radiomic recipes were developed on these examinations with B2 as the out of distribution check, and the segmentation model saw images and masks from every scanner, though never a fibrosis label (\cref{sec:data,sec:protocols}). No fold is therefore a fully independent test of the whole pipeline, the B2 fold least of all, although the comparisons do not change without it. Absolute performance on a new scanner is modest (C index 0.646), and we did not test larger staging models. Harmonization and the prevalence matched threshold were evaluated transductively, the nested fusion and the corruption simulation were designed after the primary results, and in histopathology the results for single hospitals changed with patch resolution. Most importantly, none of our conclusions has yet been tested on an external cohort. Selecting every recipe inside each outer fold and testing on such a cohort are the natural next steps.

\smallskip
\noindent
\textbf{Acknowledgments}. Shu Hu is supported by the U.S. National Science Foundation (NSF) under grant IIS-2434967. The views, opinions, and/or findings expressed are those of the author and should not be interpreted as representing the official views of NSF.

{
    \small
    \bibliographystyle{ieeenat_fullname}
    \bibliography{main}
}

\end{document}


\maketitle
\thispagestyle{empty}

This supplement gives the details and secondary results that the main paper refers to. Section and table numbers without the prefix S refer to the main paper. All intervals are 95\% bootstrap intervals computed as in the main paper (Sec.~3.3), and $p$ values are unadjusted unless a Holm decision is shown.

\section{Fusion weight and seeds}
Over the five pairs with adequate negatives, varying the forest weight from 0 to 1 keeps the mean AUC between 0.560 and 0.644, and no weight beats the deep stream alone (0.639) by more than 0.005. Over all six pairs, forest weights between 0.1 and 0.5 give 0.642 to 0.655 against 0.627, a lead that again depends on B2 S2+. Results on B2 also depend more on the seed: single seed cirrhosis AUC ranges from 0.475 to 0.726 there, so the three seed value of 0.717 should be read with this spread in mind.

For the three interventions, the single seed C index is 0.549 $\pm$ 0.009 (multiphase input), 0.615 $\pm$ 0.012 (intensity augmentation) and 0.613 $\pm$ 0.037 (adversarial scanner classifier), mean $\pm$ standard deviation over the three seeds; the main paper reports the three seed ensembles (0.562, 0.609 and 0.640).

\section{AUC per held out scanner and task}
\Cref{tab:s_auc} lists the AUC of the deep stream, the radiomic forest and their equal fusion for each held out scanner and task under leave one scanner out (LOSO) evaluation. The deep stream's interval excludes 0.5 on four of six pairs, and fusion is slightly ahead on B1 S4 ($+0.019$, 95\% CI $-0.045$ to $+0.084$). For B2 S2+, with 68 positives and five negatives, every resample reuses the same five negative cases, so both the AUC and any bootstrap interval are unreliable; we report the point estimate without an interval and draw no conclusion about model superiority from it. On the mean AUC over the five pairs with adequate negatives, the three interventions differ from the deep stream by $-0.097$ for multiphase input (95\% CI $-0.158$ to $-0.039$), $-0.011$ for intensity augmentation ($-0.043$ to $+0.020$) and $-0.013$ for the adversarial scanner classifier ($-0.042$ to $+0.016$).

\begin{table}[t]
\centering
\caption{\textbf{AUC per held out scanner and task.} Fusion uses equal weights. Negatives in parentheses; $^{*}$ deep stream interval excludes 0.5; last column: fusion minus deep with a paired, class stratified bootstrap 95\% CI. $^{\dagger}$ 68 positives and five negatives; every resample reuses the same five negatives, so no interval is reported (n.r.).}
\label{tab:s_auc}
\footnotesize
\setlength{\tabcolsep}{3pt}
\begin{tabular}{lcccc}
\toprule
Pair & Deep & Forest & Fusion & Fusion $-$ deep \\
\midrule
A, S2+ (55)   & \textbf{0.612}$^{*}$ & 0.578 & 0.601 & $-0.011$ ($-0.093$, $+0.067$) \\
A, S4 (97)    & \textbf{0.571} & 0.568 & 0.567 & $-0.004$ ($-0.079$, $+0.073$) \\
B1, S2+ (39)  & \textbf{0.615}$^{*}$ & 0.533 & 0.585 & $-0.030$ ($-0.086$, $+0.025$) \\
B1, S4 (90)   & 0.678$^{*}$ & 0.603 & \textbf{0.697} & $+0.019$ ($-0.045$, $+0.084$) \\
B2, S2+ (5)$^{\dagger}$ & 0.571 & \textbf{0.826} & 0.771 & $+0.200$ (n.r.) \\
B2, S4 (35)   & \textbf{0.717}$^{*}$ & 0.519 & 0.688 & $-0.029$ ($-0.098$, $+0.035$) \\
\midrule
Mean, six pairs & 0.627 & 0.605 & 0.651 & \\
Mean, without $^{\dagger}$ & 0.639 & 0.560 & 0.628 & \\
\bottomrule
\end{tabular}
\end{table}

\section{Radiomic learners and harmonization}
\Cref{tab:s_learners} gives the mean LOSO AUC of each classical learner under each label free adjustment. Per scanner z scoring helps the SVMs and logistic regression but moves the tree ensembles by at most 0.007, and ComBat lifts logistic regression to 0.638. Nested fusion picks one of eight radiomic pipelines by an inner LOSO over the training scanners and standardizes both risks within the test scanner, a transductive step that needs no labels.

\textbf{Learner settings.} The linear SVM uses $C = 0.1$, the RBF SVM $C = 1$, boosting runs 300 iterations at a learning rate of 0.05, every learner is class balanced, and the deep stream is trained with mixed precision. All models are implemented in PyTorch with timm, transformers, sklearn, nnUNet and inmoose, and one deep training run takes four to eight minutes on a single GPU.

\begin{table}[t]
\centering
\caption{\textbf{Radiomic learners and harmonization under leave one scanner out.} Mean AUC over the six scanner and task pairs (worst pair in parentheses) on the 198 features (ComBat on the 184 with nonzero variance on every scanner, run for logistic regression and the forest only). $\Delta$: mean minus the deep stream's mean of 0.627 (its worst pair is 0.571). Bold: best mean per column.}
\label{tab:s_learners}
\footnotesize
\setlength{\tabcolsep}{2.8pt}
\begin{tabular}{lcccc}
\toprule
Learner & Raw & Scanner z & Scanner rank & ComBat \\
\midrule
Logistic   & 0.606 (0.550) & \textbf{0.633} (0.542) & 0.602 (0.514) & \textbf{0.638} (0.509) \\
\quad $\Delta$ & $-0.021$ & $+0.006$ & $-0.025$ & $+0.011$ \\
Lin.\ SVM  & 0.570 (0.530) & 0.617 (0.527) & 0.600 (0.494) & \\
\quad $\Delta$ & $-0.057$ & $-0.010$ & $-0.027$ & \\
RBF SVM    & 0.595 (0.549) & 0.629 (0.596) & \textbf{0.618} (0.584) & \\
\quad $\Delta$ & $-0.032$ & $+0.002$ & $-0.009$ & \\
Boosting   & \textbf{0.629} (0.559) & 0.624 (0.540) & 0.616 (0.530) & \\
\quad $\Delta$ & $+0.002$ & $-0.003$ & $-0.011$ & \\
Forest     & 0.605 (0.519) & 0.612 (0.526) & 0.601 (0.541) & 0.628 (0.554) \\
\quad $\Delta$ & $-0.022$ & $-0.015$ & $-0.026$ & $+0.001$ \\
\bottomrule
\end{tabular}
\end{table}

\section{Additional series and the radiomic forest}
The forest gains from the additional series. With GED4 and shape features alone (33 features), it averages 0.561 over the six pairs and falls below chance on A for S2+ (0.402, 95\% CI 0.31 to 0.49); with all seven series (189 features, before the nine added descriptors), the average rises to 0.614. The nine descriptors, computed in voxel units or from raw GED4 intensities, rank patients below chance on all six pairs when used alone (mean AUC 0.388, with the interval excluding 0.5 on three) and cost a little in combination (0.605).

\section{Local labels, decision thresholds and four class decisions}
Local labels help the forest, most clearly on A and B1. With $k = 40$ labeled examinations from the held out scanner (Sec.~3.4), a 300 tree forest improves its AUC on A from 0.566 to 0.773 for S2+ and to 0.720 for S4, and on B1 from 0.576 to 0.693 for S4; on B2 the cirrhosis gain is smaller (0.533 to 0.595). Fixed decision thresholds transfer poorly. At the default threshold of 0.5, the deep stream's decisions largely reduce to the majority rule. For cirrhosis, its accuracy (0.618 on A, 0.391 on B1, 0.479 on B2) equals that of calling every examination noncirrhotic, with expected calibration errors of 0.193, 0.360 and 0.246, and for S2+ on A and B1 (0.650 and 0.830) it equals that of calling every examination positive. Thresholds tuned on target labels raise cirrhosis accuracy to 0.643, 0.739 and 0.712, and the prevalence matched threshold gives 0.541, 0.574 and 0.671. A model can therefore rank patients better than chance on a new scanner and still classify them poorly at a fixed cutoff. Four class decisions are also uninformative. The deep stream's most probable stage is correct for only 0.108 of the examinations on A and 0.126 on B1, where it assigns most of them to S2 or S3 (148 of 157 on A to S3, 213 of 230 on B1 to S2 or S3), and for 0.425 on B2. A four class vector built from the equally fused binaries never beats always predicting S4 (0.382 versus 0.382 on A, with a quadratic weighted kappa of 0.000; 0.530 versus 0.609 on B1; 0.493 versus 0.521 on B2).

\section{Matched seeds in distribution}
Our first analysis compared the radiomic forest with a single seed deep stream in distribution. \Cref{tab:s_id_values,tab:s_id_comp} repeat the comparison with the deep stream trained on the same three seeds as under LOSO, and \cref{tab:s_id_auc} breaks the AUCs down by scanner. With one seed, the forest appeared slightly ahead in distribution; with three seeds it is slightly behind, and in neither case is the difference significant. Under LOSO it is significantly behind, and the protocol effect is significant on both metrics.

\begin{table*}[t]\centering
\caption{\textbf{In distribution and LOSO, with matched seeds.} Pooled within scanner C index and mean AUC over the five scanner and task pairs with adequate negatives (95\% CI). In distribution: five fold cross validation stratified by scanner and stage. The seed 42 row is the single seed run of our first analysis.}\label{tab:s_id_values}
\small\setlength{\tabcolsep}{5pt}\begin{tabular}{lcccc}\toprule
 & \multicolumn{2}{c}{C index} & \multicolumn{2}{c}{Mean AUC, five pairs} \\ \cmidrule(lr){2-3}\cmidrule(lr){4-5}
Model & In distribution & LOSO & In distribution & LOSO \\ \midrule
Deep stream, three seeds & 0.750 (0.713, 0.787) & 0.646 (0.602, 0.688) & 0.755 (0.713, 0.795) & 0.639 (0.589, 0.687) \\
Deep stream, seed 42 only & 0.709 (0.667, 0.750) &  & 0.724 (0.679, 0.767) &  \\
Radiomic forest & 0.723 (0.683, 0.763) & 0.561 (0.514, 0.606) & 0.744 (0.700, 0.784) & 0.560 (0.512, 0.608) \\
Equal fusion & 0.754 (0.716, 0.790) & 0.625 (0.583, 0.668) & 0.771 (0.731, 0.810) & 0.628 (0.580, 0.673) \\
\bottomrule\end{tabular}\end{table*}

\begin{table*}[t]\centering
\caption{\textbf{Paired comparisons behind the protocol effect.} Differences (95\% CI) and two sided bootstrap $p$ values, 2,000 resamples drawn within each scanner. Protocol effect: the difference under LOSO minus the difference in distribution. Estimates are given to four decimals as computed.}\label{tab:s_id_comp}
\small\setlength{\tabcolsep}{5pt}\begin{tabular}{lcccc}\toprule
 & \multicolumn{2}{c}{C index} & \multicolumn{2}{c}{Mean AUC, five pairs} \\ \cmidrule(lr){2-3}\cmidrule(lr){4-5}
Comparison & Difference (95\% CI) & $p$ & Difference (95\% CI) & $p$ \\ \midrule
Forest $-$ deep, in distribution & $-0.0266$ ($-0.066$, $+0.014$) & 0.192 & $-0.0109$ ($-0.055$, $+0.037$) & 0.644 \\
Forest $-$ deep, in distribution, seed 42 & $+0.0146$ ($-0.032$, $+0.059$) & 0.495 & $+0.0206$ ($-0.027$, $+0.070$) & 0.408 \\
Forest $-$ deep, LOSO & $-0.0853$ ($-0.139$, $-0.027$) & 0.003 & $-0.0784$ ($-0.135$, $-0.021$) & 0.007 \\
Protocol effect, forest & $-0.0588$ ($-0.110$, $-0.007$) & 0.029 & $-0.0675$ ($-0.124$, $-0.009$) & 0.022 \\
Fusion $-$ deep, in distribution & $+0.0045$ ($-0.023$, $+0.034$) & 0.769 & $+0.0163$ ($-0.016$, $+0.051$) & 0.335 \\
Fusion $-$ deep, LOSO & $-0.0209$ ($-0.056$, $+0.018$) & 0.248 & $-0.0110$ ($-0.045$, $+0.024$) & 0.486 \\
Protocol effect, fusion & $-0.0254$ ($-0.059$, $+0.010$) & 0.141 & $-0.0273$ ($-0.069$, $+0.015$) & 0.174 \\
Deep stream, LOSO $-$ in distribution & $-0.1037$ ($-0.143$, $-0.068$) & $<0.001$ & $-0.1165$ ($-0.157$, $-0.074$) & $<0.001$ \\
Forest, LOSO $-$ in distribution & $-0.1625$ ($-0.207$, $-0.120$) & $<0.001$ & $-0.1840$ ($-0.230$, $-0.137$) & $<0.001$ \\
\bottomrule\end{tabular}\end{table*}

\begin{table*}[t]\centering
\caption{\textbf{AUC in distribution and under LOSO, per scanner and task.} Negatives in the second column. Deep, three seeds; deep, seed 42 only; forest; fusion with the three seed deep stream. Change: LOSO minus in distribution.}\label{tab:s_id_auc}
\small\setlength{\tabcolsep}{4.5pt}\begin{tabular}{llccccccccc}\toprule
 & & & \multicolumn{4}{c}{In distribution} & \multicolumn{2}{c}{LOSO} & \multicolumn{2}{c}{Change} \\ \cmidrule(lr){4-7}\cmidrule(lr){8-9}\cmidrule(lr){10-11}
Scope & Task & Neg. & Deep & Deep, seed 42 & Forest & Fusion & Deep & Forest & Deep & Forest \\ \midrule
Pooled & S2+ & 99 & 0.786 & 0.755 & 0.790 & 0.816 &  &  &  &  \\
Pooled & S4 & 222 & 0.780 & 0.753 & 0.785 & 0.803 &  &  &  &  \\
A & S2+ & 55 & 0.836 & 0.825 & 0.819 & 0.855 & 0.612 & 0.578 & $-0.224$ & $-0.242$ \\
A & S4 & 97 & 0.843 & 0.807 & 0.816 & 0.846 & 0.571 & 0.568 & $-0.272$ & $-0.248$ \\
B1 & S2+ & 39 & 0.725 & 0.647 & 0.688 & 0.724 & 0.615 & 0.533 & $-0.110$ & $-0.155$ \\
B1 & S4 & 90 & 0.785 & 0.727 & 0.762 & 0.789 & 0.678 & 0.603 & $-0.107$ & $-0.159$ \\
B2 & S2+ & 5 & 0.506 & 0.553 & 0.879 & 0.832 & 0.571 & 0.826 & $+0.065$ & $-0.053$ \\
B2 & S4 & 35 & 0.587 & 0.612 & 0.635 & 0.643 & 0.717 & 0.519 & $+0.130$ & $-0.116$ \\
\bottomrule\end{tabular}\end{table*}

\section{Comparisons without the B2 fold}
Because B2 guided the development of the deep recipe, we repeated every comparison on the A and B1 folds alone (\cref{tab:s_ab1}). No model is significantly better than the deep stream; the forest and the multiphase input remain significantly worse, and intensity augmentation becomes nominally significant, again against it ($p = 0.047$, not significant after Holm correction within family~2).

\begin{table*}[t]\centering
\caption{\textbf{Leave one scanner out without the B2 fold.} Pooled within scanner C index over the A and B1 folds, and difference from the deep stream (95\% CI, unadjusted $p$). Declared comparisons in bold.}\label{tab:s_ab1}
\small\setlength{\tabcolsep}{8pt}\begin{tabular}{lcccc}\toprule
Model & C index & $\Delta$ & 95\% CI & $p$ \\ \midrule
\textbf{Adversarial scanner classifier} & 0.646 & $+0.002$ & $-0.028$, $+0.033$ & 0.864 \\
Deep stream (GED4) & 0.644 & & & \\
Logistic regression, ComBat & 0.640 & $-0.003$ & $-0.066$, $+0.057$ & 0.941 \\
\textbf{Equal fusion} & 0.623 & $-0.020$ & $-0.060$, $+0.016$ & 0.312 \\
Fusion, scanner z & 0.608 & $-0.036$ & $-0.083$, $+0.008$ & 0.109 \\
\textbf{Frozen DINOv2 probe} & 0.605 & $-0.039$ & $-0.090$, $+0.013$ & 0.146 \\
\textbf{Intensity augmentation} & 0.603 & $-0.041$ & $-0.080$, $-0.000$ & 0.047 \\
Fusion, scanner rank & 0.602 & $-0.042$ & $-0.092$, $+0.003$ & 0.071 \\
Forest, ComBat & 0.583 & $-0.060$ & $-0.119$, $-0.005$ & 0.032 \\
Fusion (33 features), scanner z & 0.583 & $-0.061$ & $-0.110$, $-0.015$ & 0.009 \\
Fusion (33 features), scanner rank & 0.575 & $-0.068$ & $-0.119$, $-0.021$ & 0.003 \\
Fusion (33 features) & 0.571 & $-0.073$ & $-0.118$, $-0.030$ & 0.001 \\
\textbf{Multiphase deep input} & 0.568 & $-0.076$ & $-0.134$, $-0.018$ & 0.008 \\
\textbf{Radiomic forest} & 0.562 & $-0.082$ & $-0.139$, $-0.026$ & 0.004 \\
Forest, scanner z & 0.559 & $-0.085$ & $-0.145$, $-0.029$ & 0.003 \\
Forest, scanner rank & 0.553 & $-0.091$ & $-0.152$, $-0.035$ & 0.002 \\
Forest (33 features), scanner z & 0.527 & $-0.117$ & $-0.177$, $-0.058$ & $<0.001$ \\
Forest (33 features), scanner rank & 0.522 & $-0.122$ & $-0.183$, $-0.063$ & $<0.001$ \\
Forest (33 features) & 0.477 & $-0.167$ & $-0.232$, $-0.101$ & $<0.001$ \\
\bottomrule\end{tabular}\end{table*}

\section{C index per held out scanner}
\Cref{tab:s_percenter} lists the C index of every model on each held out scanner. The deep stream is above chance on all three scanners; the radiomic forest only on A.

\begin{table*}[t]\centering
\caption{\textbf{C index per held out scanner (95\% CI).} $^{*}$ interval above 0.5. Declared models in bold. ``33 features'': GED4 and shape features only.}\label{tab:s_percenter}
\small\setlength{\tabcolsep}{5pt}\begin{tabular}{lccc}\toprule
Model & A & B1 & B2 \\ \midrule
Deep stream (GED4) & 0.578 (0.51, 0.65)$^{*}$ & 0.680 (0.62, 0.74)$^{*}$ & 0.683 (0.58, 0.78)$^{*}$ \\
\textbf{Radiomic forest} & 0.575 (0.51, 0.65)$^{*}$ & 0.555 (0.49, 0.62) & 0.544 (0.41, 0.66) \\
\textbf{Equal fusion} & 0.578 (0.51, 0.65)$^{*}$ & 0.649 (0.59, 0.71)$^{*}$ & 0.654 (0.52, 0.76)$^{*}$ \\
Forest, scanner z & 0.618 (0.55, 0.68)$^{*}$ & 0.526 (0.46, 0.59) & 0.566 (0.44, 0.67) \\
Fusion, scanner z & 0.609 (0.54, 0.67)$^{*}$ & 0.607 (0.54, 0.68)$^{*}$ & 0.648 (0.53, 0.75)$^{*}$ \\
Forest, scanner rank & 0.591 (0.52, 0.66)$^{*}$ & 0.531 (0.46, 0.60) & 0.576 (0.46, 0.68) \\
Fusion, scanner rank & 0.586 (0.52, 0.65)$^{*}$ & 0.610 (0.54, 0.68)$^{*}$ & 0.657 (0.54, 0.76)$^{*}$ \\
Forest (33 features) & 0.406 (0.34, 0.48) & 0.516 (0.45, 0.58) & 0.591 (0.46, 0.72) \\
Fusion (33 features) & 0.437 (0.36, 0.51) & 0.646 (0.58, 0.71)$^{*}$ & 0.695 (0.57, 0.80)$^{*}$ \\
Forest (33 features), scanner z & 0.532 (0.46, 0.60) & 0.524 (0.46, 0.59) & 0.567 (0.45, 0.68) \\
Fusion (33 features), scanner z & 0.536 (0.46, 0.61) & 0.609 (0.54, 0.68)$^{*}$ & 0.631 (0.51, 0.73)$^{*}$ \\
Forest (33 features), scanner rank & 0.512 (0.44, 0.58) & 0.527 (0.46, 0.59) & 0.582 (0.47, 0.69) \\
Fusion (33 features), scanner rank & 0.523 (0.45, 0.59) & 0.605 (0.53, 0.67)$^{*}$ & 0.643 (0.53, 0.74)$^{*}$ \\
\textbf{Frozen DINOv2 probe} & 0.625 (0.56, 0.69)$^{*}$ & 0.594 (0.53, 0.66)$^{*}$ & 0.511 (0.41, 0.61) \\
Forest, ComBat & 0.595 (0.52, 0.66)$^{*}$ & 0.577 (0.51, 0.64)$^{*}$ & 0.587 (0.46, 0.70) \\
Logistic regression, ComBat & 0.676 (0.61, 0.74)$^{*}$ & 0.620 (0.56, 0.68)$^{*}$ & 0.679 (0.57, 0.79)$^{*}$ \\
\textbf{Multiphase deep input} & 0.482 (0.41, 0.57) & 0.616 (0.55, 0.68)$^{*}$ & 0.482 (0.37, 0.60) \\
\textbf{Intensity augmentation} & 0.567 (0.50, 0.64) & 0.622 (0.55, 0.69)$^{*}$ & 0.704 (0.61, 0.80)$^{*}$ \\
\textbf{Adversarial scanner classifier} & 0.566 (0.49, 0.64) & 0.691 (0.62, 0.75)$^{*}$ & 0.568 (0.45, 0.68) \\
\bottomrule\end{tabular}\end{table*}

\section{Scanner decodability and transfer}
\begin{figure}[t]\centering
\begin{tikzpicture}
\begin{axis}[width=\linewidth, height=5.6cm, xmin=0.845, xmax=1.005, ymin=0.47, ymax=0.66,
  xlabel={scanner AUC of the representation}, ylabel={LOSO C index},
  tick label style={font=\scriptsize, /pgf/number format/fixed, /pgf/number format/precision=2}, label style={font=\footnotesize},
  axis lines*=left, ymajorgrids, grid style={black!12},
  legend style={font=\scriptsize, draw=none, fill=none, at={(0.02,0.84)}, anchor=north west}, legend cell align=left]
\addplot[only marks, mark=*, mark size=2pt, color=cblue, mark options={draw=white, line width=0.4pt}] coordinates {(0.935,0.646) (0.882,0.562) (0.928,0.609) (0.857,0.640)};
\addlegendentry{trained deep stream}
\addplot[only marks, mark=square*, mark size=2pt, color=corange, mark options={draw=white, line width=0.4pt}] coordinates {(0.998,0.599)};
\addlegendentry{frozen DINOv2}
\addplot[only marks, mark=triangle*, mark size=2.6pt, color=caqua, mark options={draw=white, line width=0.4pt}] coordinates {(0.991,0.484) (0.999,0.530) (0.995,0.526) (0.995,0.561) (1.000,0.559) (0.999,0.554) (0.999,0.584) (0.999,0.643)};
\addlegendentry{radiomic pipelines}
\addplot[black!45, dashed, domain=0.845:1.005, samples=2, forget plot] {0.5};
\node[font=\scriptsize, text=black!60, anchor=south west] at (axis cs:0.847,0.5) {chance};
\node[font=\scriptsize, text=black!70, anchor=east, xshift=-3pt, yshift=4pt] at (axis cs:0.935,0.646) {deep (GED4)};
\node[font=\scriptsize, text=black!70, anchor=west, xshift=3pt, yshift=0pt] at (axis cs:0.882,0.562) {multiphase};
\node[font=\scriptsize, text=black!70, anchor=west, xshift=3pt, yshift=0pt] at (axis cs:0.928,0.609) {augmentation};
\node[font=\scriptsize, text=black!70, anchor=west, xshift=3pt, yshift=3pt] at (axis cs:0.857,0.640) {adversarial};
\end{axis}
\end{tikzpicture}
\caption{\textbf{Scanner decodability against transfer.} Scanner AUC against LOSO C index for the 13 representations of \cref{tab:s_dec}.}\label{fig:s_dec}
\end{figure}
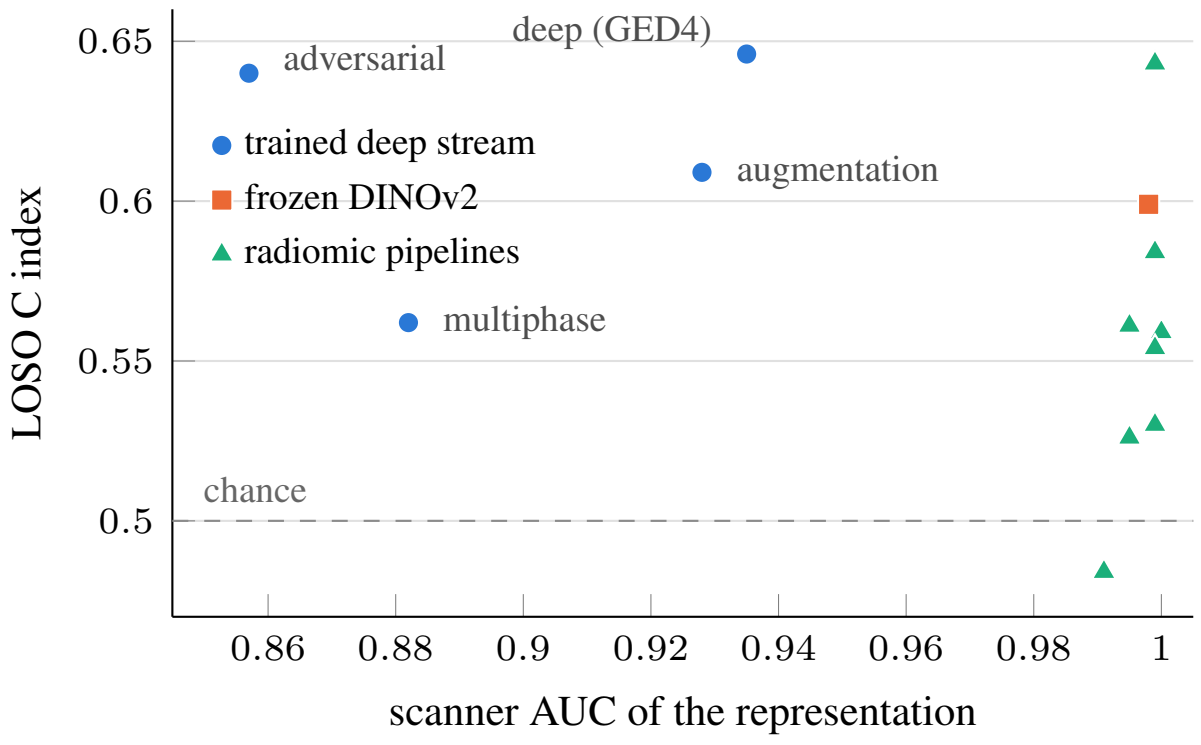

Across the 13 representations of \cref{tab:s_dec}, scanner AUC and LOSO C index are not significantly correlated (Spearman $\rho = -0.37$, $p = 0.21$; \cref{fig:s_dec}). Among the four trained deep embeddings, the correlation is positive ($\rho = +0.40$, four points): the two interventions that removed the most scanner information transferred no better (adversary) or worse (multiphase input). The scanner accuracies of these embeddings are 0.839 (baseline), 0.752 (multiphase input), 0.826 (intensity augmentation) and 0.713 (adversary), against a majority rate of 0.500.

\begin{table}[t]\centering
\caption{\textbf{Scanner decodability and transfer.} Scanner AUC: one versus rest AUC, macro averaged, of a 600 tree forest predicting the scanner from the representation (five fold cross validation). Deep embeddings come from the models trained with B2 held out (seed 42); radiomic values are from our original analysis.}\label{tab:s_dec}
\footnotesize\setlength{\tabcolsep}{4pt}\begin{tabular}{lcc}\toprule
Representation & Scanner AUC & LOSO C index \\ \midrule
Deep stream (GED4) & 0.935 & 0.646 \\
Multiphase deep input & 0.882 & 0.562 \\
Intensity augmentation & 0.928 & 0.609 \\
Adversarial scanner classifier & 0.857 & 0.640 \\
Frozen DINOv2 & 0.998 & 0.599 \\
Forest (33 features) & 0.991 & 0.484 \\
Forest (33 features), scanner z & 0.999 & 0.530 \\
Forest (33 features), scanner rank & 0.995 & 0.526 \\
Radiomic forest & 0.995 & 0.561 \\
Forest, scanner z & 1.000 & 0.559 \\
Forest, scanner rank & 0.999 & 0.554 \\
Forest, ComBat & 0.999 & 0.584 \\
Logistic regression, ComBat & 0.999 & 0.643 \\
\bottomrule\end{tabular}\end{table}

\section{Corruption simulation}
\begin{figure*}[t]\centering
\begin{tikzpicture}
\begin{axis}[width=0.5\textwidth, height=5.2cm, xmin=6, xmax=54, ymin=-0.1, ymax=0.025, xtick={10,25,50}, title={examinations chosen at random}, title style={font=\footnotesize},
  xlabel={\% of the scanner's examinations corrupted}, ylabel={change in C index}, 
  tick label style={font=\scriptsize, /pgf/number format/fixed, /pgf/number format/precision=2}, label style={font=\footnotesize},
  axis lines*=left, ymajorgrids, grid style={black!12}, legend style={font=\scriptsize, draw=none, fill=none, at={(0.02,0.02)}, anchor=south west}, legend cell align=left]
\addplot[black!45, line width=0.5pt, domain=6:54, samples=2, forget plot] {0};
\addplot[color=cblue, line width=1.1pt, mark=*, mark size=2pt, mark options={solid, draw=white, line width=0.3pt}] coordinates {(10,-0.004) (25,-0.017) (50,-0.031)};
\addlegendentry{three series zeroed}
\addplot[color=cblue, line width=0.7pt, dashed, forget plot] coordinates {(10,0.003) (25,-0.003) (50,0.004)};
\addplot[color=corange, line width=1.1pt, mark=square*, mark size=2pt, mark options={solid, draw=white, line width=0.3pt}] coordinates {(10,-0.001) (25,-0.017) (50,-0.041)};
\addlegendentry{all seven series zeroed}
\addplot[color=corange, line width=0.7pt, dashed, forget plot] coordinates {(10,-0.002) (25,-0.008) (50,-0.004)};
\addplot[color=caqua, line width=1.1pt, mark=triangle*, mark size=2pt, mark options={solid, draw=white, line width=0.3pt}] coordinates {(10,-0.004) (25,-0.015) (50,-0.010)};
\addlegendentry{shift of 1 SD}
\addplot[color=caqua, line width=0.7pt, dashed, forget plot] coordinates {(10,0.000) (25,0.002) (50,-0.002)};
\addplot[color=cyellow, line width=1.1pt, mark=diamond*, mark size=2pt, mark options={solid, draw=white, line width=0.3pt}] coordinates {(10,-0.001) (25,-0.013) (50,-0.013)};
\addlegendentry{shift of 2 SD}
\addplot[color=cyellow, line width=0.7pt, dashed, forget plot] coordinates {(10,0.000) (25,0.000) (50,-0.003)};
\addlegendimage{black!70, line width=1.1pt}\addlegendentry{held out scanner (solid)}
\addlegendimage{black!70, line width=0.7pt, dashed}\addlegendentry{pooled folds (dashed)}
\end{axis}
\end{tikzpicture}\hfill
\begin{tikzpicture}
\begin{axis}[width=0.5\textwidth, height=5.2cm, xmin=6, xmax=54, ymin=-0.1, ymax=0.025, xtick={10,25,50}, title={examinations chosen in proportion to stage}, title style={font=\footnotesize},
  xlabel={\% of the scanner's examinations corrupted}, 
  tick label style={font=\scriptsize, /pgf/number format/fixed, /pgf/number format/precision=2}, label style={font=\footnotesize},
  axis lines*=left, ymajorgrids, grid style={black!12}]
\addplot[black!45, line width=0.5pt, domain=6:54, samples=2, forget plot] {0};
\addplot[color=cblue, line width=1.1pt, mark=*, mark size=2pt, mark options={solid, draw=white, line width=0.3pt}] coordinates {(10,-0.013) (25,-0.033) (50,-0.038)};
\addplot[color=cblue, line width=0.7pt, dashed, forget plot] coordinates {(10,0.002) (25,0.006) (50,0.009)};
\addplot[color=corange, line width=1.1pt, mark=square*, mark size=2pt, mark options={solid, draw=white, line width=0.3pt}] coordinates {(10,-0.021) (25,-0.061) (50,-0.078)};
\addplot[color=corange, line width=0.7pt, dashed, forget plot] coordinates {(10,0.006) (25,0.002) (50,0.009)};
\addplot[color=caqua, line width=1.1pt, mark=triangle*, mark size=2pt, mark options={solid, draw=white, line width=0.3pt}] coordinates {(10,-0.010) (25,-0.033) (50,-0.052)};
\addplot[color=caqua, line width=0.7pt, dashed, forget plot] coordinates {(10,0.005) (25,0.002) (50,0.013)};
\addplot[color=cyellow, line width=1.1pt, mark=diamond*, mark size=2pt, mark options={solid, draw=white, line width=0.3pt}] coordinates {(10,-0.019) (25,-0.052) (50,-0.090)};
\addplot[color=cyellow, line width=0.7pt, dashed, forget plot] coordinates {(10,0.006) (25,0.004) (50,0.009)};
\end{axis}
\end{tikzpicture}
\caption{\textbf{Injected errors under pooled and held out evaluation.} Mean change in C index over three scanners and three draws; solid: corrupted scanner held out; dashed: pooled folds, all scanners.}\label{fig:s_corr}
\end{figure*}
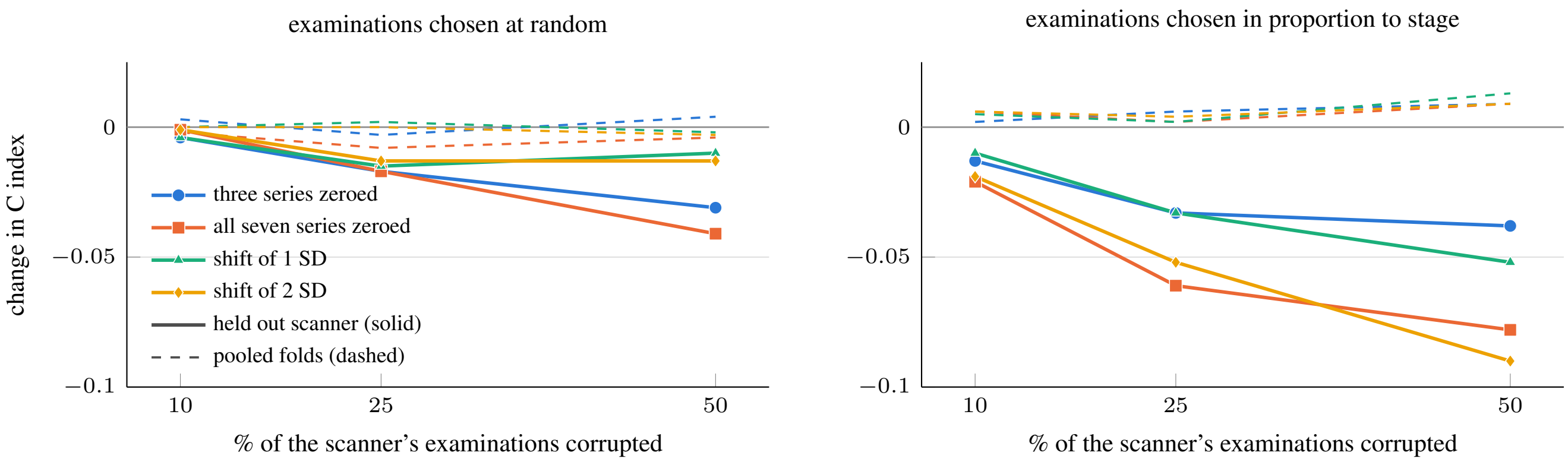

\Cref{tab:s_corr} and \cref{fig:s_corr} give every setting of the simulation described in Sec.~3.4. Under pooled folds, the C index over all scanners moves by at most 0.013, whereas with the corrupted scanner held out it falls in every stage linked setting, more so as the corrupted share grows. When the corruption follows stage, the pooled C index measured on the corrupted scanner itself can even rise (up to $+0.046$), because the corruption then carries stage information that pooled folds let the forest learn.

\begin{table*}[t]\centering
\caption{\textbf{Corruption simulation.} Change in C index of a 300 tree forest from the uncorrupted features, mean (standard deviation) over the three scanners and three random draws. Pooled, all: pooled cross validation, C index over all scanners. Pooled, target: the same predictions, C index on the corrupted scanner only. Held out: the corrupted scanner left out of training. Uncorrupted reference: pooled 0.727 over all scanners; on A, B1 and B2, pooled 0.779, 0.711, 0.602 and held out 0.572, 0.547, 0.553.}\label{tab:s_corr}
\small\setlength{\tabcolsep}{6pt}\begin{tabular}{llcccc}\toprule
Corruption & Examinations & Share & Pooled, all & Pooled, target & Held out \\ \midrule
All seven zeroed & random & 10\% & $-0.002$ (0.006) & $-0.019$ (0.033) & $-0.001$ (0.009) \\
All seven zeroed & random & 25\% & $-0.008$ (0.006) & $-0.042$ (0.035) & $-0.017$ (0.017) \\
All seven zeroed & random & 50\% & $-0.004$ (0.010) & $-0.048$ (0.057) & $-0.041$ (0.040) \\
All seven zeroed & by stage & 10\% & 0.006 (0.008) & 0.013 (0.032) & $-0.021$ (0.031) \\
All seven zeroed & by stage & 25\% & 0.002 (0.015) & $-0.006$ (0.039) & $-0.061$ (0.045) \\
All seven zeroed & by stage & 50\% & 0.009 (0.009) & 0.010 (0.045) & $-0.078$ (0.075) \\
\midrule
Three series zeroed & random & 10\% & 0.003 (0.004) & $-0.007$ (0.018) & $-0.004$ (0.010) \\
Three series zeroed & random & 25\% & $-0.003$ (0.004) & $-0.025$ (0.028) & $-0.017$ (0.011) \\
Three series zeroed & random & 50\% & 0.004 (0.007) & $-0.010$ (0.028) & $-0.031$ (0.019) \\
Three series zeroed & by stage & 10\% & 0.002 (0.005) & 0.001 (0.017) & $-0.013$ (0.028) \\
Three series zeroed & by stage & 25\% & 0.006 (0.009) & 0.003 (0.036) & $-0.033$ (0.028) \\
Three series zeroed & by stage & 50\% & 0.009 (0.006) & 0.046 (0.048) & $-0.038$ (0.066) \\
\midrule
Shift, 1 SD & random & 10\% & 0.000 (0.002) & $-0.013$ (0.022) & $-0.004$ (0.012) \\
Shift, 1 SD & random & 25\% & 0.002 (0.005) & $-0.022$ (0.028) & $-0.015$ (0.033) \\
Shift, 1 SD & random & 50\% & $-0.002$ (0.006) & $-0.032$ (0.043) & $-0.010$ (0.034) \\
Shift, 1 SD & by stage & 10\% & 0.005 (0.005) & 0.006 (0.019) & $-0.010$ (0.023) \\
Shift, 1 SD & by stage & 25\% & 0.002 (0.005) & $-0.016$ (0.032) & $-0.033$ (0.041) \\
Shift, 1 SD & by stage & 50\% & 0.013 (0.009) & 0.031 (0.045) & $-0.052$ (0.078) \\
\midrule
Shift, 2 SD & random & 10\% & 0.000 (0.009) & $-0.016$ (0.029) & $-0.001$ (0.025) \\
Shift, 2 SD & random & 25\% & 0.000 (0.010) & $-0.022$ (0.030) & $-0.013$ (0.035) \\
Shift, 2 SD & random & 50\% & $-0.003$ (0.007) & $-0.045$ (0.039) & $-0.013$ (0.027) \\
Shift, 2 SD & by stage & 10\% & 0.006 (0.005) & $-0.003$ (0.034) & $-0.019$ (0.021) \\
Shift, 2 SD & by stage & 25\% & 0.004 (0.005) & $-0.006$ (0.046) & $-0.052$ (0.041) \\
Shift, 2 SD & by stage & 50\% & 0.009 (0.014) & 0.003 (0.050) & $-0.090$ (0.038) \\
\bottomrule\end{tabular}\end{table*}

\section{MiniCAMELYON17}
MiniCAMELYON17 uses only lymph nodes 0 and 1 from the CAMELYON17 training set, about 121 slides, according to its dataset card. Its $512\times512$ patches were sampled from tissue at 0.5\,$\mu$m per pixel (about $20\times$), with a quota for each slide instead of uniform sampling, so tumor patches are far more common than in the original slides. A patch is tumor if an annotated metastasis overlaps its central third, and normal if its central third has no metastasis and is at least 20\% tissue; patches with metastasis only outside the central third were dropped, as in WILDS. The card lists five centers, while the release has six center identifiers, which we keep as released (\cref{tab:s_comp}). The main analysis resizes each patch to $224\times224$, and the native resolution check takes the central $224\times224$ crop.

\begin{table}[t]\centering
\caption{\textbf{MiniCAMELYON17 composition.} Center identifiers as released. Centers 4 and 5 contain no tumor patches and are excluded.}\label{tab:s_comp}
\footnotesize\setlength{\tabcolsep}{4pt}\begin{tabular}{cccccc}\toprule
Center & Patients & Slides & Patches & Tumor & Tumor (\%) \\ \midrule
0 & 19 & 32 & 10,484 & 1,500 & 14.3 \\
1 & 15 & 23 & 6,639 & 31 & 0.5 \\
2 & 15 & 23 & 8,100 & 1,200 & 14.8 \\
3 & 2 & 2 & 630 & 30 & 4.8 \\
4 & 2 & 2 & 600 & 0 & 0.0 \\
5 & 1 & 1 & 300 & 0 & 0.0 \\
\midrule
0 to 3 & 51 & 80 & 25,853 & 2,761 & 10.7 \\
\bottomrule\end{tabular}\end{table}

\section{Additional histopathology results}
\Cref{tab:s_hosp,tab:s_dec_path,tab:s_nn,tab:s_dir,tab:s_native} give the per hospital, decodability, nearest neighbor, tumor direction and native resolution results summarized in the main paper. Fitting the harmonization on all patches before forming the folds takes the hospital AUC further below 0.5 than fitting it on the training folds only; the latter matches a probe without features, so the values below 0.5 come from how patient grouped folds shift each hospital's share between training and test, not from hospital information. On native resolution crops, the effects reported in the main paper hold, while the worst hospital and the order of the encoders on single hospitals change (\cref{tab:s_hosp,tab:s_native}).

\begin{table}[t]\centering
\caption{\textbf{LOHO tumor AUC per held out hospital,} resized patches (main analysis) and native resolution central crops, no harmonization.}\label{tab:s_hosp}
\footnotesize\setlength{\tabcolsep}{2.4pt}\begin{tabular}{ccccccc}\toprule
 & \multicolumn{3}{c}{Resized} & \multicolumn{3}{c}{Native} \\ \cmidrule(lr){2-4}\cmidrule(lr){5-7}
Hospital & DINOv2 & Phikon & Phikon-v2 & DINOv2 & Phikon & Phikon-v2 \\ \midrule
0 & 0.838 & 0.981 & 0.972 & 0.849 & 0.960 & 0.979 \\
1 & 0.908 & 0.956 & 0.923 & 0.940 & 0.905 & 0.928 \\
2 & 0.923 & 0.972 & 0.974 & 0.910 & 0.859 & 0.983 \\
3 & 0.838 & 0.998 & 0.956 & 0.688 & 0.999 & 0.953 \\
\bottomrule\end{tabular}\end{table}

\begin{table}[t]\centering
\caption{\textbf{Hospital decodability} (AUC / accuracy) of the logistic probe under patient grouped five fold cross validation. Harmonization fitted on all patches before the folds are formed, or on the training folds only. A probe without features reaches 0.312 / 0.286; the majority rate is 0.41.}\label{tab:s_dec_path}
\footnotesize\setlength{\tabcolsep}{3pt}\begin{tabular}{lccc}\toprule
Embedding & DINOv2 & Phikon & Phikon-v2 \\ \midrule
Raw & 0.978 / 0.933 & 0.999 / 0.988 & 0.999 / 0.981 \\
z scoring, all patches & 0.287 / 0.253 & 0.179 / 0.145 & 0.182 / 0.146 \\
ComBat, all patches & 0.287 / 0.253 & 0.179 / 0.145 & 0.182 / 0.147 \\
z scoring, training folds & 0.311 / 0.286 & 0.309 / 0.286 & 0.314 / 0.286 \\
\bottomrule\end{tabular}\end{table}

\begin{table}[t]\centering
\caption{\textbf{Nearest neighbor of each patch} (cosine similarity, resized patches): share from the same slide, patient, hospital and label. Chance: expected share for a random neighbor.}\label{tab:s_nn}
\footnotesize\setlength{\tabcolsep}{4pt}\begin{tabular}{lcccc}\toprule
 & Slide & Patient & Hospital & Label \\ \midrule
DINOv2 & 0.498 & 0.506 & 0.890 & 0.989 \\
Phikon & 0.906 & 0.907 & 0.999 & 0.999 \\
Phikon-v2 & 0.888 & 0.889 & 0.998 & 0.999 \\
Chance & 0.014 & 0.022 & 0.329 & 0.809 \\
\bottomrule\end{tabular}\end{table}

\begin{table}[t]\centering
\caption{\textbf{Tumor direction across patients} (11 slides from 11 patients with at least 20 tumor and 20 normal patches). Direction of a slide: mean tumor minus mean normal standardized embedding. Cosine: mean cosine between the directions of two slides. AUC: mean over ordered pairs of slides from different patients of the AUC of one slide's direction on the other slide's patches (95\% CI, 2,000 resamples of slides).}\label{tab:s_dir}
\footnotesize\setlength{\tabcolsep}{3pt}\begin{tabular}{lccc}\toprule
 & Cosine & AUC, resized & AUC, native \\ \midrule
DINOv2 & 0.139 & 0.608 (0.465, 0.760) & 0.614 (0.470, 0.760) \\
Phikon & 0.188 & 0.842 (0.665, 0.961) & 0.852 (0.687, 0.963) \\
Phikon-v2 & 0.159 & 0.791 (0.609, 0.933) & 0.804 (0.624, 0.942) \\
\bottomrule\end{tabular}\end{table}

\begin{table*}[t]\centering
\caption{\textbf{Resized patches against native resolution central crops.} Tumor AUC of the same probe under each protocol.}\label{tab:s_native}
\small\setlength{\tabcolsep}{6pt}\begin{tabular}{lcccccc}\toprule
 & \multicolumn{2}{c}{DINOv2} & \multicolumn{2}{c}{Phikon} & \multicolumn{2}{c}{Phikon-v2} \\ \cmidrule(lr){2-3}\cmidrule(lr){4-5}\cmidrule(lr){6-7}
Protocol & Resized & Native & Resized & Native & Resized & Native \\ \midrule
Random patch split & 0.998 & 0.999 & 1.000 & 1.000 & 1.000 & 1.000 \\
Folds grouped by patient & 0.849 & 0.863 & 0.990 & 0.984 & 0.985 & 0.990 \\
LOHO, mean & 0.877 & 0.847 & 0.977 & 0.931 & 0.957 & 0.961 \\
LOHO, worst hospital & 0.838 & 0.688 & 0.956 & 0.859 & 0.923 & 0.928 \\
LOHO, hospital z scoring & 0.872 & 0.828 & 0.929 & 0.915 & 0.925 & 0.929 \\
LOHO, ComBat & 0.876 & 0.828 & 0.933 & 0.915 & 0.921 & 0.929 \\
\bottomrule\end{tabular}\end{table*}